\documentclass{article}

\usepackage{PRIMEarxiv}
\usepackage[utf8]{inputenc}
\usepackage[T1]{fontenc}
\usepackage{hyperref}
\usepackage{url}
\usepackage{graphicx}
\usepackage{amsmath,amssymb}
\usepackage{amsfonts}
\usepackage{booktabs}
\usepackage{multirow}
\usepackage{xcolor}
\usepackage{array}
\usepackage{microtype}
\usepackage{fancyhdr}

\graphicspath{{media/}}

\begin{document}
\title{SAGA: A Robust Self-Attention and Goal-Aware Anchor-based Planner for Safe UAV Autonomous Navigation}
\author{
Junhao Wei$^{1,2}$,
Yanxiao Li$^{1}$,
Dexing Yao$^{1}$,
Yifu Zhao$^{1}$,
Haochen Li$^{1}$,
Qibin He$^{1}$,
Baili Lu$^{3}$,
Xiaofan Zou$^{4}$,\\
Dingcheng Yang$^{5}$,
Sio-Kei Im$^{6}$,
Yapeng Wang$^{1}$,
Xu Yang$^{1,*}$\\
\small $^{1}$Faculty of Applied Sciences, Macao Polytechnic University, Macao 999078, China\\
\small $^{2}$Pazhou Lab (Huangpu), Guangzhou 510555, China\\
\small $^{3}$College of Animal Science and Technology, Zhongkai University of Agriculture and Engineering, Guangzhou 510225, China\\
\small $^{4}$School of Mechanical and Electrical Engineering and Automation, Shanghai University, Shanghai 200444, China\\
\small $^{5}$Information Engineering School, Nanchang University, Nanchang 330031, China\\
\small $^{6}$Macao Polytechnic University, Macao 999078, China
}
\maketitle

\begin{abstract}
Agile unmanned aerial vehicle (UAV) navigation in cluttered environments demands a planning architecture that is both computationally efficient and structurally expressive enough to reason over multiple feasible motions. This paper presents SAGA, a robust self-attention and goal-aware anchor-based planner for safe UAV autonomous navigation. SAGA formulates local planning as a one-stage joint regression-and-ranking problem over a fixed lattice of motion anchors. Given a depth image and a body-frame motion state, the planner predicts refined terminal states and planning scores for all anchors in a single forward pass, after which the best candidate is decoded into a dynamically feasible trajectory. The key idea of SAGA is to transform anchor-aligned features into geometry-aware tokens and perform cross-anchor global reasoning with self-attention. To preserve directional structure in the token space, we further introduce a polar positional encoding derived from anchor yaw and pitch. In addition, a goal-aware modulation module injects velocity, acceleration, and target information into the token representation before final score prediction. Experiments in cluttered pillar-map environments under maximum speed settings of 2.0, 3.0, and 4.0~m/s show that SAGA consistently achieves a 100\% success rate, while YOPO drops from 90.91\% to 62.50\%, Ego-planner from 71.43\% to 52.63\%, and Fast-planner from 52.63\% to 38.46\%. Under the 4.0~m/s maximum speed setting, SAGA also improves average safety from 1.9843~m to 2.3888~m and minimum safety from 0.4390~m to 0.7576~m over YOPO, while reducing total flight time from 40.4631~s to 27.4901~s. The comparison with SAGA w/o PPE further shows that explicit polar positional encoding is critical for stable cross-anchor reasoning and safe passage selection in cluttered scenes.

\end{abstract}


\section{Introduction}
Autonomous navigation for quadrotors in cluttered environments remains challenging because safe flight must be achieved under strict real-time constraints, partial onboard perception, and highly multimodal local decision spaces. In scenes such as pillar forests, narrow corridors, and obstacle clusters, the vehicle often encounters several dynamically feasible motion choices at each replanning step, while only a subset of them remain simultaneously safe, goal-directed, and smooth. As a result, a practical navigation system must not only react quickly, but must also compare multiple candidate motions in a structured and reliable manner.

Classical local planning pipelines typically decompose the problem into mapping, search, trajectory optimization, and low-level control. This decomposition provides clear geometric interpretation, but it also increases computational latency and repeatedly solves similar subproblems in every control cycle. In contrast, learning-based one-stage planners offer much lower inference latency by directly predicting candidate motions from perception. However, many of these models still evaluate motion candidates largely independently, which limits their ability to resolve globally coupled local decisions. In cluttered environments, the preference for one candidate often depends on the availability and relative quality of its neighboring candidates. This coupling is especially important near bottlenecks, where a UAV must effectively negotiate passage selection among several competing anchors.

Recent UAV navigation research has developed along several representative directions. In 2021, Zhou \emph{et al.} proposed Ego-planner, an ESDF-free gradient-based local planner that avoids maintaining a full signed distance field and instead formulates collision penalties through a collision-guided path representation, thereby improving computational efficiency while preserving trajectory smoothness and robustness \cite{EGO}. In 2025, Xu \emph{et al.} proposed NavRL, a deep reinforcement learning framework based on PPO for safe UAV navigation in dynamic environments, where a learned policy is combined with a lightweight safety shield to reduce collision risk and improve sim-to-real transfer \cite{NAVRL}. Also in 2025, Lu \emph{et al.} proposed YOPO, a learning-based one-stage planner that integrates perception, local search, and trajectory optimization into a single network by regressing primitive-wise trajectory refinements and scores directly from depth observations \cite{YOPO}. These works demonstrate the strengths of optimization-based, reinforcement learning-based, and one-stage learning-based paradigms, respectively. However, the one-stage anchor-based line still suffers from a key limitation: while multiple motion candidates are predicted in parallel, they are not globally reasoned about with sufficiently explicit geometric structure.

\begin{figure}[t]
    \centering
    \includegraphics[width=\textwidth]{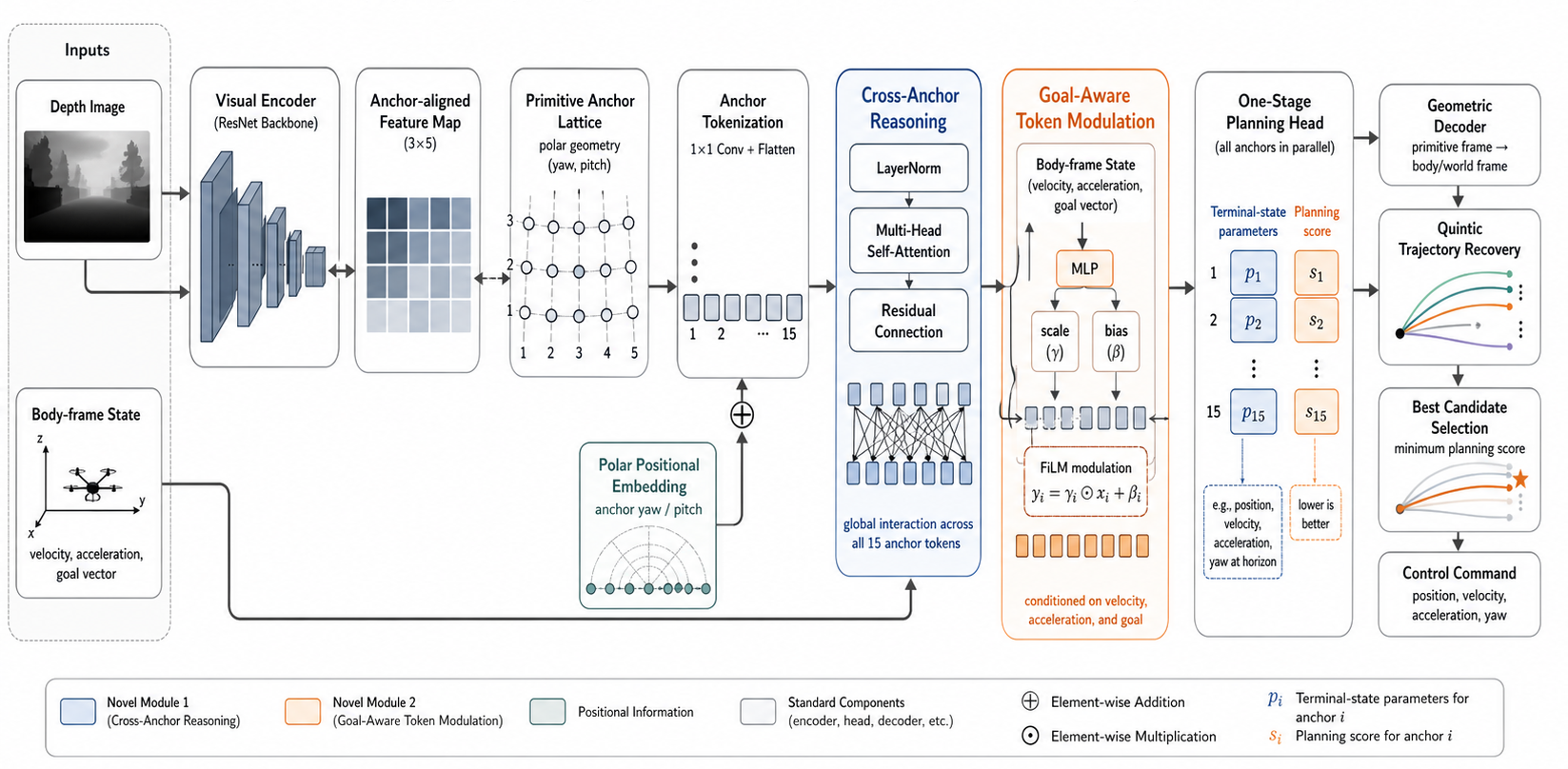}
    \caption{Overall architecture of the proposed SAGA framework. Given a depth image and a body-frame motion state, SAGA encodes anchor-aligned visual features, performs cross-anchor reasoning with self-attention and polar positional encoding, applies goal-aware token modulation, and predicts terminal-state refinements and planning scores for all motion anchors in one forward pass. The best candidate is then decoded into a dynamically feasible trajectory for execution.}
    \label{yuanli}
\end{figure}

This paper proposes \emph{SAGA}, a \textbf{S}elf-\textbf{A}ttention and \textbf{G}oal-\textbf{A}ware \textbf{A}nchor-based planner for safe UAV autonomous navigation. SAGA preserves the efficiency and structure of anchor-based one-stage planning, but upgrades the anchor evaluation process from independent scoring to global cross-anchor reasoning. Specifically, the planner first encodes a depth image into an anchor-aligned feature map, then converts anchor cells into a set of tokens, and finally performs self-attention over all anchor tokens. To preserve the directional geometry of the anchor lattice, SAGA introduces a polar positional encoding derived from the yaw and pitch of each anchor. The resulting token features are further modulated by the current body-frame velocity, acceleration, and goal vector, enabling the planner to select candidates that are not only collision-aware but also dynamically and task consistent. The overall pipeline of SAGA is illustrated in Figure~\ref{yuanli}.

SAGA is designed as a structured one-stage planning framework rather than an end-to-end black-box control policy. The anchor lattice remains explicit, the decoding process remains analytic, and the final selected motion is converted into a quintic polynomial trajectory for execution. This combination of explicit motion structure and learned candidate reasoning makes SAGA suitable for safe and fast receding-horizon flight.

The contributions of this paper are summarized as follows:
\begin{itemize}
\item We propose SAGA, a one-stage anchor-based UAV planner that jointly predicts terminal-state refinements and planning scores over a fixed motion lattice.
\item We introduce a cross-anchor self-attention mechanism that performs global reasoning over motion anchors while preserving one-stage planning efficiency.
\item We design a polar positional encoding and a goal-aware token modulation scheme that inject geometric and task-dependent priors into anchor-token reasoning.
\item We validate SAGA against learning-based and optimization-based baselines, including YOPO \cite{YOPO}, Ego-planner \cite{EGO}, and Fast-planner \cite{FAST}, and further show through ablation that polar positional encoding plays a critical role in robust anchor reasoning.
\end{itemize}

\section{The Proposed SAGA}

\subsection{Problem Formulation}
At every replanning step, the UAV receives a depth image
\begin{equation}
\mathbf{D} \in \mathbb{R}^{1 \times H \times W}
\end{equation}
and a body-frame motion state
\begin{equation}
\mathbf{o} = [\mathbf{v}^{b}, \mathbf{a}^{b}, \mathbf{g}^{b}] \in \mathbb{R}^{9},
\end{equation}
where $\mathbf{v}^{b}$ and $\mathbf{a}^{b}$ denote the current velocity and acceleration in the body frame, and $\mathbf{g}^{b}$ is the body-frame goal vector.

Instead of directly regressing a control action, SAGA defines a fixed set of motion anchors and predicts a refined terminal state together with a planning score for every anchor. The candidate with the lowest predicted cost is then decoded and converted into a dynamically feasible polynomial trajectory.

\subsection{Anchor-based Planning Space}
\begin{figure}[htbp]
    \centering
    \includegraphics[width=0.5\textwidth]{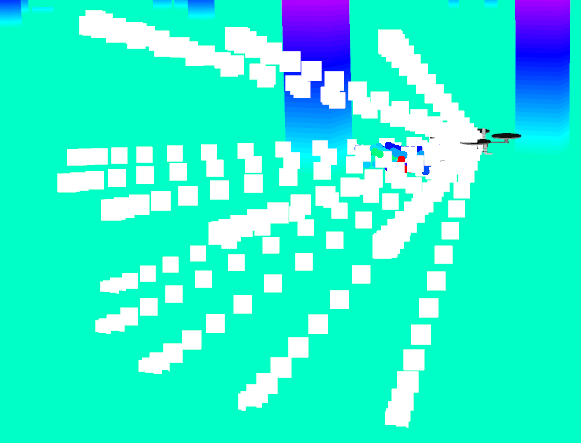}
    \caption{Visualization of the motion anchor lattice used in SAGA. The white cubic markers denote the predefined primitive anchors distributed over the local field of view. Instead of regressing trajectories directly in the continuous action space, SAGA predicts geometric residuals relative to these anchors, allowing the anchor lattice to deform adaptively according to the observed environment.}
    \label{pri6}
\end{figure}

\begin{figure}[htbp]
    \centering
    \includegraphics[width=0.5\textwidth]{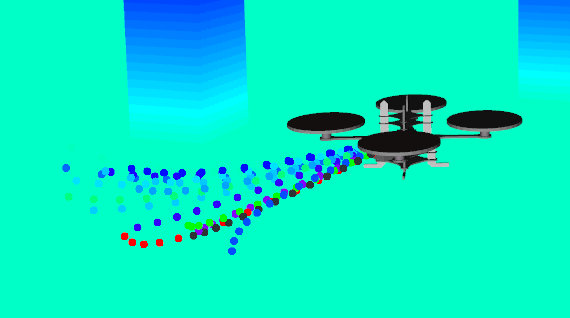}
    \caption{Visualization of candidate trajectories generated from the primitive anchor lattice. These candidates provide diverse coverage of the local solution space and are evaluated jointly by SAGA to select the final executed trajectory.}
    \label{pri5}
\end{figure}

To address the intrinsic multi-modality of local UAV trajectory generation, SAGA adopts a motion primitive lattice, as illustrated in Figure~\ref{pri6} and Figure~\ref{pri5} \cite{TGK}. The lattice provides a discrete set of anchor trajectories that cover the local solution space from different directions. Rather than directly regressing a single absolute trajectory in continuous space, SAGA predicts terminal-state refinements relative to these primitive anchors and evaluates all candidates in parallel. This design converts trajectory generation into a structured anchor-based prediction problem, which improves candidate diversity and stabilizes one-stage planning in cluttered environments.

Let $\mathcal{A}=\{\mathbf{a}_i\}_{i=1}^{N}$ denote a fixed lattice of motion anchors. Each anchor is parameterized by its nominal yaw and pitch:
\begin{equation}
\mathbf{a}_i = [\alpha_i, \beta_i]^\top.
\end{equation}
In our implementation, the anchor lattice is arranged on a $3 \times 5$ grid, yielding $N=15$ anchor hypotheses. These anchors define a structured local planning space that covers candidate directions in front of the vehicle.

For each anchor, the planner predicts a 9-dimensional terminal-state parameter vector
\begin{equation}
\hat{\mathbf{u}}_i = [\delta \alpha_i, \delta \beta_i, r_i, \mathbf{v}_i^p, \mathbf{a}_i^p],
\end{equation}
where $\delta \alpha_i$ and $\delta \beta_i$ refine the angular direction of the anchor, $r_i$ controls the radial extent, and $\mathbf{v}_i^p,\mathbf{a}_i^p$ denote terminal velocity and acceleration in the primitive frame. In addition, the planner predicts a scalar planning score $\hat{c}_i$ for each anchor.

\subsection{Visual Encoding and Anchor Tokenization}
The depth image is first encoded by a convolutional backbone:
\begin{equation}
\mathbf{F} = E_\theta(\mathbf{D}), \qquad \mathbf{F}\in\mathbb{R}^{C \times 3 \times 5},
\end{equation}
where the output resolution is aligned with the anchor lattice. A pointwise projection lifts the feature dimension to a hidden token space:
\begin{equation}
\mathbf{H} = \phi(\mathbf{F}), \qquad \mathbf{H}\in\mathbb{R}^{C_h \times 3 \times 5},
\end{equation}
with hidden dimension $C_h=256$.

The projected feature map is then flattened into a set of anchor tokens:
\begin{equation}
\mathbf{Z}^{(0)} = [\mathbf{z}_1,\mathbf{z}_2,\dots,\mathbf{z}_N]^\top \in \mathbb{R}^{N \times C_h}.
\end{equation}
Each token corresponds to one anchor and therefore inherits a clear geometric interpretation.

\subsection{Cross-Anchor Self-Attention}
In cluttered navigation, the value of a candidate motion cannot be assessed independently of its neighboring candidates. To model this interdependence, SAGA performs self-attention over the anchor-token set.

The core attention block is
\begin{equation}
\mathbf{Z}^{(1)} =
\tilde{\mathbf{Z}} +
\mathrm{MHA}\big(
\mathrm{LN}(\tilde{\mathbf{Z}}),
\mathrm{LN}(\tilde{\mathbf{Z}}),
\mathrm{LN}(\tilde{\mathbf{Z}})
\big),
\end{equation}
where $\mathrm{LN}(\cdot)$ is Layer Normalization, $\mathrm{MHA}(\cdot)$ is multi-head self-attention, and $\tilde{\mathbf{Z}}$ denotes the geometry-enhanced token sequence defined below.

This block enables every candidate anchor to reason about all other candidates jointly. Rather than relying on isolated local evidence, the planner can compare multiple passage options globally and suppress candidates that are only locally attractive.

\subsection{Polar Positional Encoding}
Self-attention alone treats tokens as an unordered set unless geometry is explicitly injected. For anchor-based planning, this is insufficient because the planner must know whether one anchor is to the left, right, above, or below another anchor. SAGA therefore introduces a polar positional encoding using the anchor yaw-pitch coordinates:
\begin{equation}
\mathbf{e}_i = W_{pe}\mathbf{a}_i + \mathbf{b}_{pe}.
\end{equation}
The token input to the attention layer becomes
\begin{equation}
\tilde{\mathbf{z}}_i = \mathbf{z}_i + \mathbf{e}_i.
\end{equation}
The resulting sequence
\begin{equation}
\tilde{\mathbf{Z}} = [\tilde{\mathbf{z}}_1,\dots,\tilde{\mathbf{z}}_N]^\top
\end{equation}
preserves anchor geometry while allowing global token interaction.

This design is important because anchor reasoning is not permutation invariant. The planner should distinguish, for example, whether two high-score anchors lie on opposite sides of an obstacle or belong to the same side of a corridor opening.

\subsection{Goal-Aware Token Modulation}
After global cross-anchor reasoning, the planner still needs to condition candidate ranking on the current dynamics and target direction. SAGA therefore modulates the token representation using the body-frame state vector $\mathbf{o}$.

We map the state vector through a lightweight multilayer perceptron:
\begin{equation}
[\boldsymbol{\gamma}, \boldsymbol{\beta}] = \mathrm{MLP}(\mathbf{o}),
\end{equation}
where $\boldsymbol{\gamma}, \boldsymbol{\beta}\in\mathbb{R}^{C_h}$. The attention-updated tokens are then modulated as
\begin{equation}
\mathbf{Z}^{(2)} = \mathrm{LN}(\mathbf{Z}^{(1)}) \odot \left(1+\tanh(\boldsymbol{\gamma})\right) + \boldsymbol{\beta}.
\end{equation}
This goal-aware modulation biases candidate selection toward motions that are compatible with both the goal direction and the instantaneous UAV dynamics.

\subsection{One-Stage Prediction Head}
The modulated token set is reshaped back into the anchor lattice and processed by a final pointwise prediction layer:
\begin{equation}
\mathbf{Y}=\psi(\mathbf{Z}^{(2)}) \in \mathbb{R}^{10 \times 3 \times 5}.
\end{equation}
For each anchor, the first nine channels define a normalized terminal-state prediction and the last channel defines a nonnegative planning score:
\begin{equation}
\hat{\mathbf{u}}_i = \tanh(\mathbf{Y}_{i}^{1:9}), \qquad
\hat{c}_i = \mathrm{softplus}(\mathbf{Y}_{i}^{10}).
\end{equation}

\subsection{Geometric Decoding}
The predicted terminal-state parameters are analytically decoded into body-frame terminal states. Let $\Delta \alpha_{\max}$ and $\Delta \beta_{\max}$ denote the maximum angular refinement range, and let $R_{\max}$ denote the radial range. The refined terminal position of anchor $i$ is
\begin{equation}
\mathbf{p}_i^b =
r_i
\begin{bmatrix}
\cos(\beta_i+\delta\beta_i)\cos(\alpha_i+\delta\alpha_i)\\
\cos(\beta_i+\delta\beta_i)\sin(\alpha_i+\delta\alpha_i)\\
\sin(\beta_i+\delta\beta_i)
\end{bmatrix}.
\end{equation}
The primitive-frame terminal velocity and acceleration are transformed to the body frame using the anchor rotation matrix $R_{bp,i}$:
\begin{equation}
\mathbf{v}_i^b = R_{bp,i}\mathbf{v}_i^p,\qquad
\mathbf{a}_i^b = R_{bp,i}\mathbf{a}_i^p.
\end{equation}

\subsection{Polynomial Trajectory Recovery}
Given the current state and a decoded terminal state, SAGA generates a smooth local trajectory by solving for a fifth-order polynomial in each axis:
\begin{equation}
q_k(t) = \sum_{m=0}^{5} c_{k,m} t^m, \qquad k\in\{x,y,z\},
\end{equation}
subject to initial and terminal constraints on position, velocity, and acceleration. The candidate associated with the minimum predicted score is selected as the executed trajectory.

\subsection{Training Objective}
For each anchor candidate, SAGA evaluates a structured trajectory cost composed of smoothness, safety, goal alignment, and acceleration regularization:
\begin{equation}
J_i = \lambda_s J_i^{\text{smooth}} + \lambda_c J_i^{\text{safe}} + \lambda_g J_i^{\text{goal}} + \lambda_a J_i^{\text{acc}}.
\end{equation}
The trajectory supervision objective is the mean cost over all anchors:
\begin{equation}
\mathcal{L}_{traj} = \frac{1}{N}\sum_{i=1}^{N}J_i.
\end{equation}
The score branch is trained to regress the same structured cost target:
\begin{equation}
\mathcal{L}_{score} = \frac{1}{N}\sum_{i=1}^{N}\mathrm{SmoothL1}(\hat{c}_i, J_i).
\end{equation}
The final loss is
\begin{equation}
\mathcal{L} = \lambda_{traj}\mathcal{L}_{traj} + \lambda_{score}\mathcal{L}_{score}.
\end{equation}
This formulation preserves explicit planning semantics while enabling end-to-end optimization of a one-stage planning architecture.

\section{Experiments}
\subsection{Experimental Setup}
We evaluate SAGA in cluttered randomized pillar-map environments using a receding-horizon simulation framework. All methods are tested under the same controller interface, target specification, and evaluation protocol. The UAV receives depth observations and must reach a goal position inside a dense obstacle field while maintaining dynamic feasibility and collision avoidance.

The compared methods are:
\begin{itemize}
\item \textbf{YOPO}: a one-stage anchor-based learning planner used as the primary learning baseline.
\item \textbf{Ego-planner}: a representative optimization-based trajectory planner.
\item \textbf{Fast-planner}: a strong classical planning baseline with explicit search-and-optimization structure.
\item \textbf{SAGA w/o PPE}: our targeted ablation that removes polar positional encoding (PPE) while keeping the anchor-token reasoning backbone.
\item \textbf{SAGA}: the full method.
\end{itemize}

We evaluate all methods under three maximum speed settings: 2.0, 3.0, and 4.0~m/s. We report success rate ($Success~Rate$), total flight time ($Time~Consumption$), trajectory length ($Traj.~Length$), average obstacle clearance ($Avg.~Safety$), minimum obstacle clearance ($Min.~Safety$), and smoothness cost ($Smoothness$). The safety metrics are computed from the executed trajectory by querying the nearest obstacle point for each sampled trajectory state. $Avg.\ Safety$ is the mean nearest-obstacle distance, while $Min.\ Safety$ is the minimum nearest-obstacle distance along the trajectory. $Smoothness$ is defined as the time integral of squared jerk, i.e., $\int \|\mathbf{j}(t)\|^2 dt$, where $\mathbf{j}(t)$ is the third-order derivative of position.

\subsection{Main Results and Analysis}
Table~\ref{tab:performance_comparison} presents the quantitative comparison across three maximum speed settings. Overall, SAGA achieves the best balance between robustness, safety, and trajectory quality. Compared with YOPO, SAGA performs stronger global candidate reasoning; compared with Ego-planner and Fast-planner, SAGA maintains much higher reliability while preserving competitive trajectory efficiency.

\begin{table}[htbp]
\centering
\caption{Quantitative comparison of robust flight performance. '$\uparrow$' indicates that larger
  values are better, while '$\downarrow$' indicates that smaller values are preferred. Here, Avg.
  safety and Min. safety denote the mean and minimum distance from the executed trajectory to the
  nearest obstacle points, respectively, and Smoothness denotes the integrated squared jerk cost
  along the trajectory. Best values are highlighted in bold.}
\label{tab:performance_comparison}
\resizebox{\textwidth}{!}{%
\begin{tabular}{llcccccc}
\toprule
\multirow{2}{*}{\textbf{Velocity}} & \multirow{2}{*}{\textbf{Metrics}} & \multicolumn{5}{c}{\textbf{Algorithms}} \\
\cmidrule(lr){3-7}
& & \textbf{YOPO} & \textbf{Ego-planner} & \textbf{Fast-planner} & \textbf{SAGA w/o PPE} & \textbf{SAGA} \\
\midrule
\multirow{6}{*}{2.0 m/s}
  & Time Consumption ($s$)$\downarrow$     & 64.0781 & 46.3180 & \textbf{41.0446} & 58.4651 &
  53.8621 \\
  & Traj. Length ($m$)$\downarrow$   & 86.6784 & 73.1619 & \textbf{71.9794} & 81.7755 & 78.5237 \\
  & Avg. Safety ($m$)$\uparrow$     & 2.1609  & 1.1645  & 1.5816  & 2.1557  & \textbf{2.4599} \\
  & Min. Safety ($m$)$\uparrow$     & 0.3971  & 0.3197  & 0.4067  & 0.7364  & \textbf{0.8049} \\
  & Smoothness ($m^2/s^5$)$\downarrow$     & 3.9302  & 412376.5579 & 31514.1764 & \textbf{2.1500} &
  2.2706 \\
  & Success Rate$\uparrow$    & 90.91\% & 71.43\% & 52.63\% & \textbf{100.00\%} & \textbf{100.00\%}
  \\
  \midrule
  \multirow{6}{*}{3.0 m/s}
  & Time Consumption ($s$)$\downarrow$     & 49.4341 & 26.5820 & \textbf{20.4670} & 39.8301 & 32.3001 \\
  & Traj. Length ($m$)$\downarrow$   & 86.5572 & 74.3906 & \textbf{74.1371} & 82.2966 & 76.4802 \\
  & Avg. Safety ($m$)$\uparrow$    & 2.0559  & 1.3049  & 1.7133  & 2.1254  & \textbf{2.3549} \\
  & Min. Safety ($m$)$\uparrow$    & 0.4253  & 0.3995  & 0.4551  & 0.6773  & \textbf{0.8936} \\
  & Smoothness ($m^2/s^5$)$\downarrow$     & 23.4272 & 479573.4814 & 113320.0796 & 15.6493 &
  \textbf{15.5506} \\
  & Success Rate$\uparrow$   & 83.33\% & 66.67\% & 47.62\% & \textbf{100.00\%} & \textbf{100.00\%} \\
  \midrule
  \multirow{6}{*}{4.0 m/s}
  & Time Consumption ($s$)$\downarrow$      & 40.4631 & 19.0290 & \textbf{15.6908} & 31.6292 & 27.4901 \\
  & Traj. Length ($m$)$\downarrow$   & 85.4210 & 74.6566 & \textbf{74.1624} & 83.1104 & 78.8110 \\
  & Avg. Safety ($m$)$\uparrow$    & 1.9843  & 1.4382  & 1.6069  & 2.0845  & \textbf{2.3888} \\
  & Min. Safety ($m$)$\uparrow$    & 0.4390  & 0.4116  & 0.4331  & 0.6709  & \textbf{0.7576} \\
  & Smoothness ($m^2/s^5$)$\downarrow$     & 112.0421 & 1472632.8490 & 191842.7380 &
  \textbf{64.5070} & 75.1067 \\
  & Success Rate$\uparrow$   & 62.50\% & 52.63\% & 38.46\% & 95.24\% & \textbf{100.00\%} \\
\bottomrule
\end{tabular}
}
\end{table}

\begin{figure}[htbp]
\centering
\includegraphics[width=0.7\textwidth]{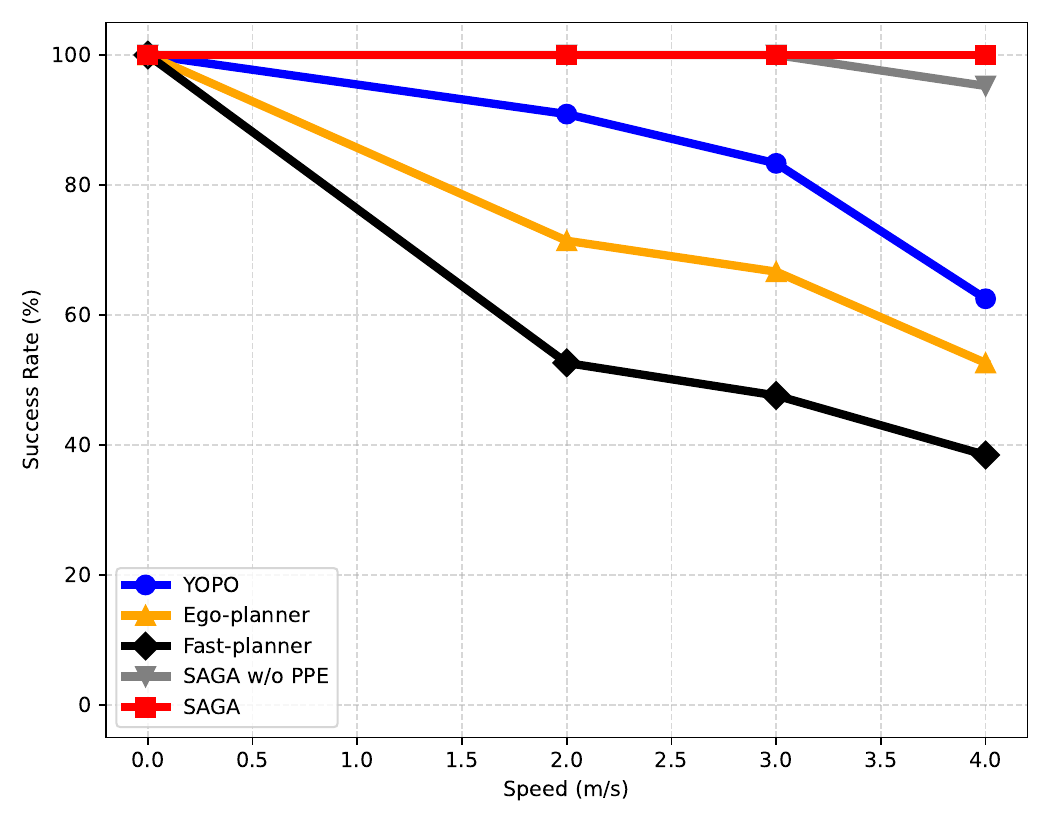}
\caption{Success rate of different planners under increasing maximum flight speed. SAGA consistently maintains the highest robustness across all tested velocity settings, while the performance of YOPO, Ego-planner, and Fast-planner degrades more significantly as the task becomes more aggressive.}
\label{sr}
\end{figure}

Several observations can be drawn from Table~\ref{tab:performance_comparison}. First, SAGA is the only method that maintains a 100\% success rate at all three tested maximum speed settings. YOPO degrades from 90.91\% at 2.0~m/s to 62.50\% at 4.0~m/s, Ego-planner decreases from 71.43\% to 52.63\%, and Fast-planner drops from 52.63\% to 38.46\%. This result indicates that SAGA is substantially more robust as the navigation task becomes more aggressive. This trend is also consistent with the success-rate curve shown in Figure~\ref{sr}.

Second, SAGA consistently provides the strongest safety performance. At 2.0~m/s, SAGA achieves the highest average safety margin of 2.4599~m and the highest minimum safety margin of 0.8049~m. At 3.0~m/s, its minimum safety margin further rises to 0.8936~m, which is more than double that of YOPO (0.4253~m) and clearly larger than those of Ego-planner and Fast-planner. At 4.0~m/s, SAGA still preserves the best average safety (2.3888~m) and minimum safety (0.7576~m), showing that the proposed planner is not only more likely to succeed, but also keeps larger clearance to obstacles throughout flight.

Third, the comparison with \emph{SAGA w/o PPE} directly verifies the effectiveness of the polar positional encoding module. The ablated version already improves significantly over YOPO, which shows that the anchor-token formulation and global reasoning are beneficial. However, the full SAGA model is consistently better. At 2.0~m/s, introducing polar positional encoding improves average safety from 2.1557~m to 2.4599~m and minimum safety from 0.7364~m to 0.8049~m, while also reducing the total time from 58.4651~s to 53.8621~s. At 3.0~m/s, SAGA reduces total time from 39.8301~s to 32.3001~s and improves minimum safety from 0.6773~m to 0.8936~m. At 4.0~m/s, the ablated version drops to 95.24\% success, whereas the full SAGA still maintains 100\% success and stronger clearance margins. This shows that attention alone is not enough; explicit polar geometry is necessary to stabilize directional reasoning among anchor candidates.

Finally, the table highlights the trade-off between pure aggressiveness and reliable safe flight. Ego-planner and Fast-planner often achieve shorter traversal time and shorter path length, but they do so at the cost of much lower success rates and substantially weaker safety margins. SAGA is therefore not the fastest planner in raw traversal time, but it offers the most favorable overall balance among success, safety, and trajectory quality, which is more valuable in safety-critical autonomous UAV navigation.

\subsection{Qualitative Analysis}
\begin{figure}[htbp]
    \centering
    \includegraphics[width=0.6\textwidth]{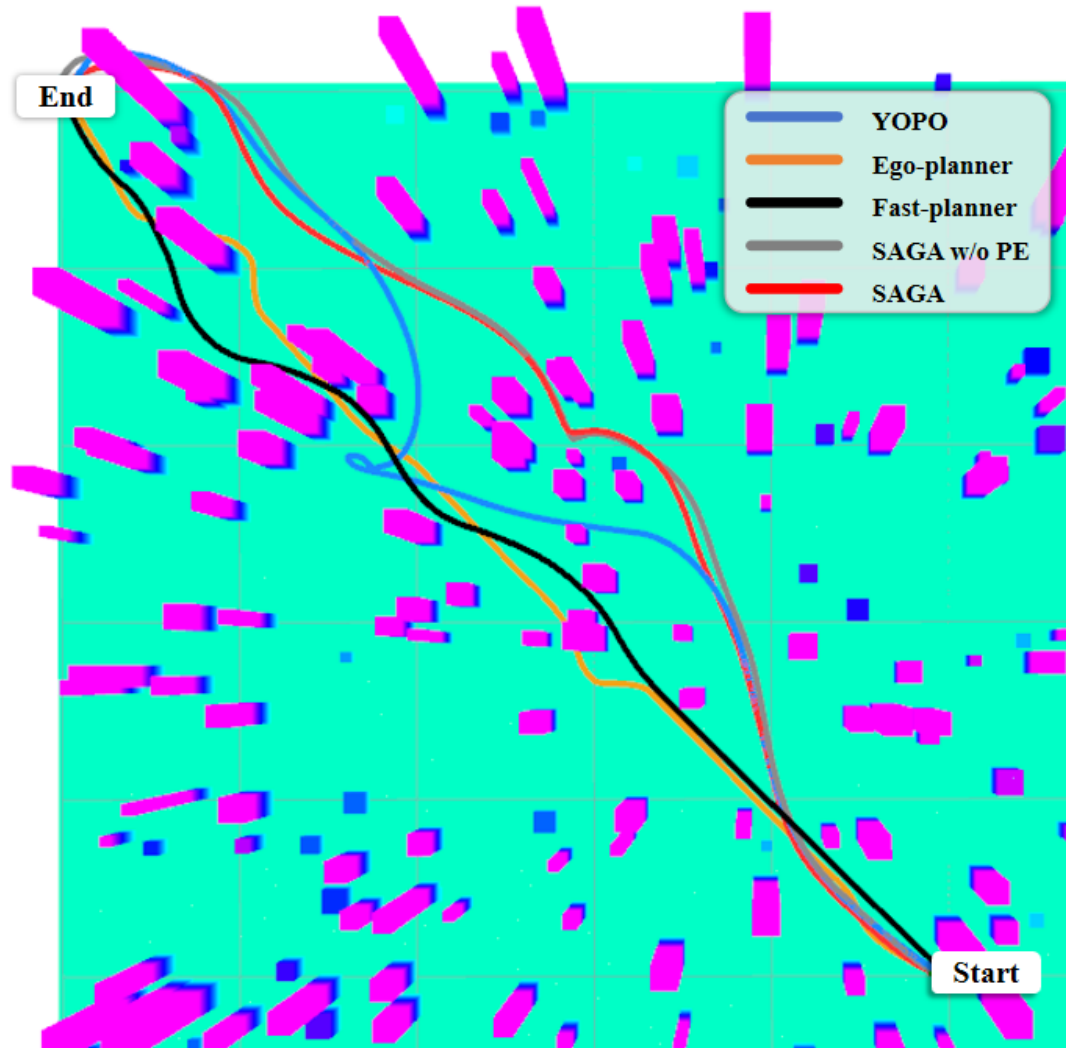}
    \caption{Qualitative comparison of flight trajectories generated by different planners at a maximum speed of $2~\text{m/s}$. The global map is visualized only for analysis and is not provided to the planners during online navigation.}
    \label{v2}
\end{figure}

\begin{figure}[htbp]
    \centering
    \includegraphics[width=0.6\textwidth]{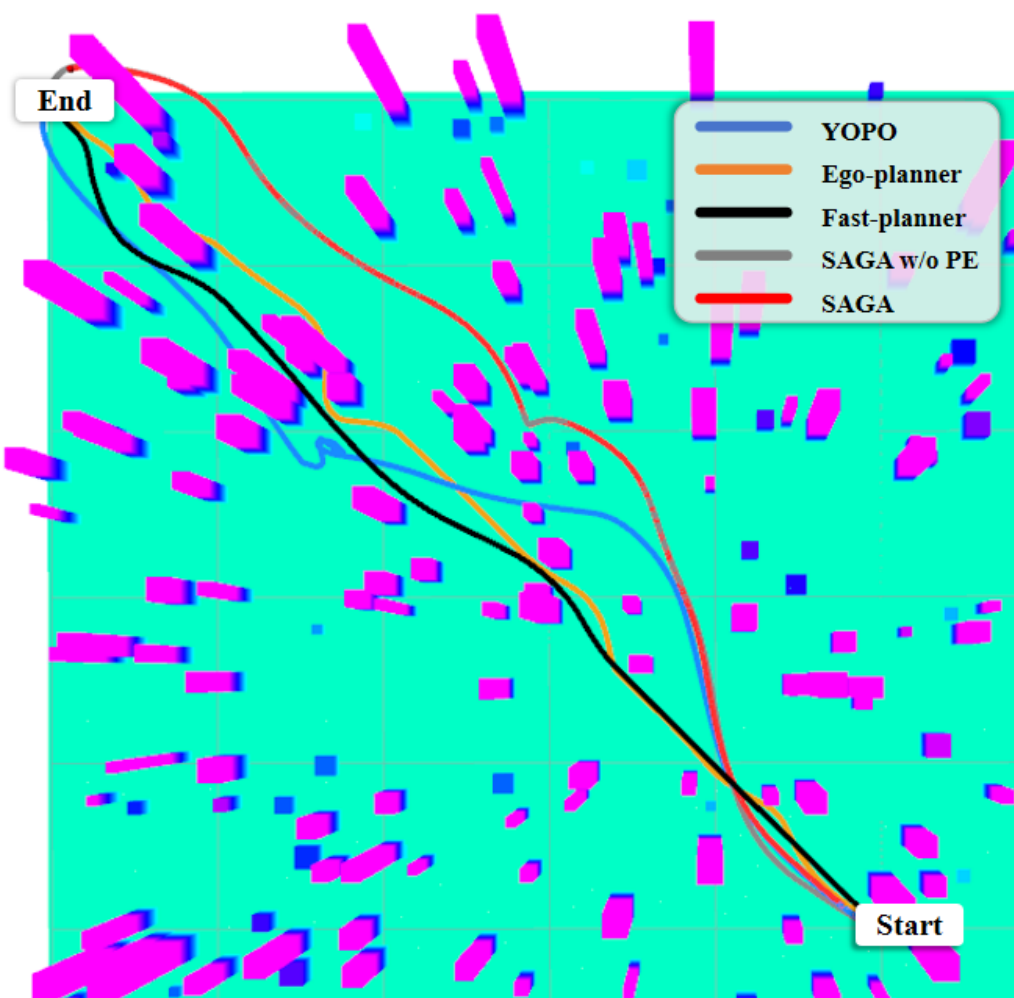}
    \caption{Qualitative comparison of flight trajectories generated by different planners at a maximum speed of $3~\text{m/s}$. SAGA produces safer and more coherent trajectories in cluttered regions than the competing methods.}
    \label{v3}
\end{figure}

\begin{figure}[htbp]
    \centering
    \includegraphics[width=0.6\textwidth]{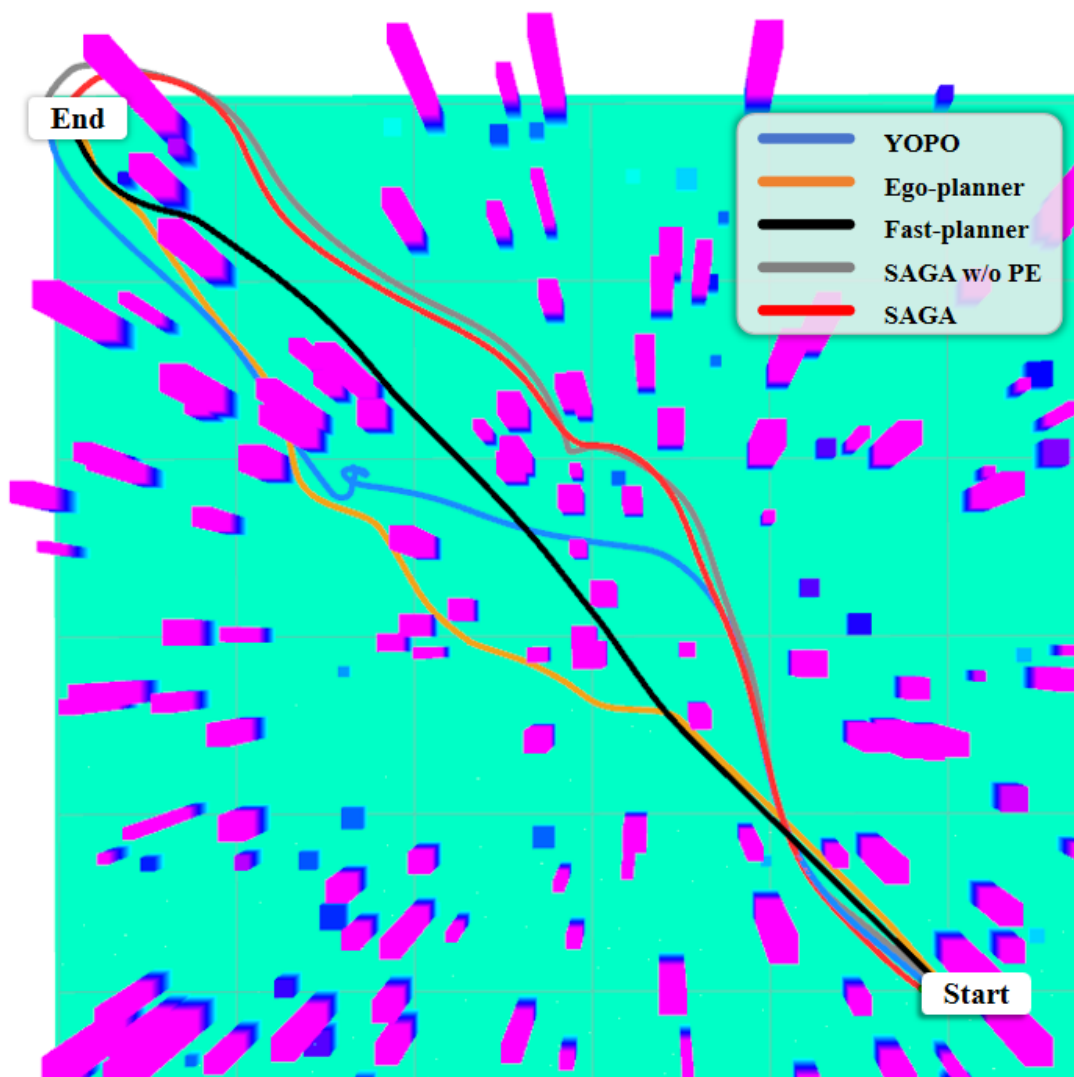}
    \caption{Qualitative comparison of flight trajectories generated by different planners at a maximum speed of $4~\text{m/s}$. As the flight speed increases, SAGA remains more robust and maintains larger obstacle clearance than the baselines.}
    \label{v4}
\end{figure}

Qualitative trajectory comparisons under the 2.0, 3.0, and 4.0~m/s maximum speed settings are shown in Figure~\ref{v2}, Figure~\ref{v3}, and Figure~\ref{v4}, respectively. These visualizations show that SAGA produces more coherent anchor selection in pillar-map bottlenecks. Compared with the one-stage baseline, SAGA is less prone to oscillatory switching among neighboring anchors and exhibits more stable motion commitment near narrow passages. Compared with the no-positional-encoding ablation, the full model better preserves directional consistency in anchor selection, particularly when several candidate gaps appear visually similar in local depth observations.

\section{Discussion}
The results support the main design hypothesis of this paper: the central limitation of one-stage anchor planning is not the anchor representation itself, but the lack of structured global interaction among anchor candidates. By converting anchor cells into tokens and enabling cross-anchor reasoning, SAGA strengthens local planning decisions without sacrificing one-stage efficiency.

The targeted ablation also clarifies an important point. Global interaction alone is not sufficient for reliable planning in cluttered environments. Since anchor candidates correspond to directional motion hypotheses, explicit geometric encoding remains necessary. Polar positional encoding gives the planner a directional prior over the anchor lattice, allowing self-attention to reason in a geometry-aware rather than permutation-agnostic manner.

SAGA still has limitations. First, it remains a local receding-horizon planner and therefore cannot fully resolve globally deceptive environments without additional long-range guidance. Second, the anchor lattice is fixed, which simplifies inference but limits adaptive spatial coverage. Third, the current planner relies on depth-only perception and does not explicitly exploit semantic scene information.

Future work will investigate how SAGA can be extended beyond single-UAV local navigation into broader aerial autonomy problems, including UAV swarms \cite{EGOSWARM}, autonomous exploration \cite{FUEL}, target tracking \cite{TRACK}, and cooperative pursuit-and-encirclement tasks \cite{pursuit}. These settings require stronger multi-agent coordination, task allocation, and long-horizon decision consistency, which suggests combining the current anchor-token reasoning pipeline with optimization-driven coordination layers. Another promising direction is to couple SAGA with adaptive optimization modules for anchor selection, cooperative trajectory refinement, and resource-aware deployment, building on our prior studies in UAV path planning and optimization \cite{wei_wiopt_2024} \cite{wei_sensors_2025} \cite{ASKSSA} \cite{MRBMO}. We also plan to explore lightweight global guidance so that token-based local reasoning can benefit from topological structure and mission-level intent.

\section{Conclusion}
This paper presented SAGA, a robust self-attention and goal-aware anchor-based planner for safe UAV autonomous navigation. SAGA formulates local planning as a one-stage anchor-token prediction problem and introduces three key ingredients: cross-anchor self-attention, polar positional encoding, and goal-aware token modulation. Together, these components enable geometry-aware global candidate reasoning and more stable target-consistent trajectory selection in cluttered environments.

Experiments against YOPO, Ego-planner, Fast-planner, and a targeted no-positional-encoding ablation demonstrate that SAGA improves planning robustness and safety while preserving the computational advantages of one-stage planning. More broadly, SAGA shows that anchor-based planning can be significantly strengthened by explicit token-level geometric reasoning, offering a practical and scalable direction for safe autonomous UAV navigation.

\section*{Acknowledgments}
The supports provided by Macao Polytechnic University (RP/FCA-01/2025) and Macao Science and Technology Development Fund (FDCT-MOST: 0018/2025/AMJ) enabled us to conduct data collection, analysis, and interpretation, as well as cover expenses related to research materials and participant recruitment. MPU and FDCT investment in our work have significantly contributed to the quality and impact of our research findings.

\section*{Competing Interests}
The authors have no competing interests to declare that are relevant to the content of this article.

\end{document}